\documentclass[twocolumn]{article}

\usepackage[pagenumbers]{arxiv}

\usepackage[a4paper, margin=20mm]{geometry}

\usepackage[linesnumbered,ruled,vlined]{algorithm2e}
\usepackage{graphicx}
\usepackage{amsmath}
\usepackage{amssymb}
\usepackage{booktabs}
\usepackage{hyphenat}
\usepackage[inline]{enumitem}
\usepackage[breaklinks,colorlinks]{hyperref}
\usepackage{multirow}

\usepackage[capitalize]{cleveref}

\newcommand{\best}[1]{\textbf{\textcolor{red}{#1}}}
\newcommand{\sbest}[1]{\textbf{\textcolor{blue}{#1}}}

\begin{document}

\title{Post-Train Adaptive U-Net for Image Segmentation}

\author{Kostiantyn Khabarlak \\
Dnipro University of Technology, Ukraine\\
{\tt\small habarlack@gmail.com}
}

\maketitle

\begin{abstract}Typical neural network architectures used for image segmentation cannot be changed without further training. This is quite limiting as the network might not only be executed on a powerful server, but also on a mobile or edge device. Adaptive neural networks offer a solution to the problem by allowing certain adaptivity after the training process is complete. In this work for the first time, we apply Post-Train Adaptive (PTA) approach to the task of image segmentation. We introduce U-Net+PTA neural network, which can be trained once, and then adapted to different device performance categories. The two key components of the approach are PTA blocks and PTA-sampling training strategy. The post-train configuration can be done at runtime on any inference device including mobile. Also, the PTA approach has allowed to improve image segmentation Dice score on the CamVid dataset. The final trained model can be switched at runtime between 6 PTA configurations, which differ by inference time and quality. Importantly, all of the configurations have better quality than the original U-Net (No PTA) model.
\end{abstract}

\keywords{Adaptive Convolutional Neural Networks, Image Segmentation, Inference Speed, Mobile Computing, Edge Computing, Computer Vision.}

\blfootnote{
  This is the peer reviewed version of the following article: \nohyphens{K. Khabarlak, ``Post-Train Adaptive U-Net for Image Segmentation,'' Information Technology: Computer Science, Software Engineering and Cyber Security, no. 2, pp. 73--78, 2022, doi: 10.32782/IT/2022-2-8.}
}

\section{Introduction}

Many fields benefit from fast and accurate image segmentation. Convolutional neural networks show the best accuracy solving the task. Applications include medical imaging~\cite{UNet}, autonomous driving~\cite{CamVid}, satellite imaging, \etc. Typical neural network architectures used for image segmentation are expected to be fully configured before the training procedure starts. To change the network architecture additional training steps are required. This is quite limiting as the network might not only be executed on a powerful server, but also on a mobile or edge device~\cite{FaceDetectionMobile,FastLandmark}. Training separate networks for each device category is quite inefficient. Ideally, the network configuration change should be performed dynamically at runtime.

Adaptive neural networks offer a solution to the problem by allowing certain adaptivity after the training process is complete. Successful approaches to building adaptive neural networks have been proposed for Recurrent Neural Networks in~\cite{ACT-RNN}, Convolutional Neural Networks in~\cite{PTA,ACT-ResNet,FasterMaml}. In particular, we see the Post-Train Adaptive approach proposed in~\cite{PTA} as an easy and effective way for the neural network adaptivity. Still the approach was only applied to the image classification task.

In this work we present U-Net+PTA network for the image segmentation task. We base upon U-Net~\cite{UNet} architecture with MobileNetV2~\cite{MobileNetV2} backbone. To enable post-train adaptivity of the network, we apply the Post-Train Adaptive approach from~\cite{PTA}.

To summarize, our main contributions are as follows:
\begin{enumerate}
  \item We introduce U-Net+PTA neural network, which can be trained once, and then adapted to devices of different performance categories.
  \item We demonstrate that U-Net+PTA not only improves inference speed over the U-Net, but also shows better Dice\textsubscript{score} on the CamVid~\cite{CamVid} dataset.
\end{enumerate}

\section{Literature Overview}

To solve the segmentation task with high quality, the input image should be considered at multiple scales. This can be done through feature pyramid network~\cite{FPN}, U-Net-like architecture~\cite{UNet} or feature exchange between multiple scales~\cite{HRNet}. Such architectures are computationally intensive. In the meantime, segmentation algorithms are often required to run on desktop as well as mobile devices, while current architectures are mostly suited for desktop applications only.

Typically, neural network architectures made to solve the segmentation task are configured before the training process starts. Different backbones can be used in the segmentation models to change their quality and inference speed, like ResNet~\cite{ResNet}, MobileNetV2~\cite{MobileNetV2}, MobileNetV3~\cite{MobileNetV3}, SENet~\cite{SENet} or others. Still, these backbones cannot be additionally configured after the training process is complete, which limits their applicability to devices with different computational resources.

Dynamic neural networks is a promising research direction~\cite{ACT-RNN,PTA,ACT-ResNet,FasterMaml}. The goal is to allow the neural network to change its architecture depending on expected inference time or input data complexity. However, in many cases additional adaptivity comes at increased computational cost. Thus, inference time might not be smaller on average than that of a conventional static neural network. In contrast, in~\cite{PTA} Post-Train Adaptive approach was presented, which via a simple MobileNetV2 modification has allowed to reduce actual inference time. Importantly, the approach allows to reconfigure the neural network after the training process is complete. But, to the best of our knowledge, the approach has only been applied to the image classification task, specifically face anti-spoofing. In this work we adapt the Post-Train Adaptive approach to the task of image segmentation.

\section{Materials and Methods}\label{sec:materials-methods}

\begin{figure*}
  \centering
  \includegraphics[width=0.8\linewidth]{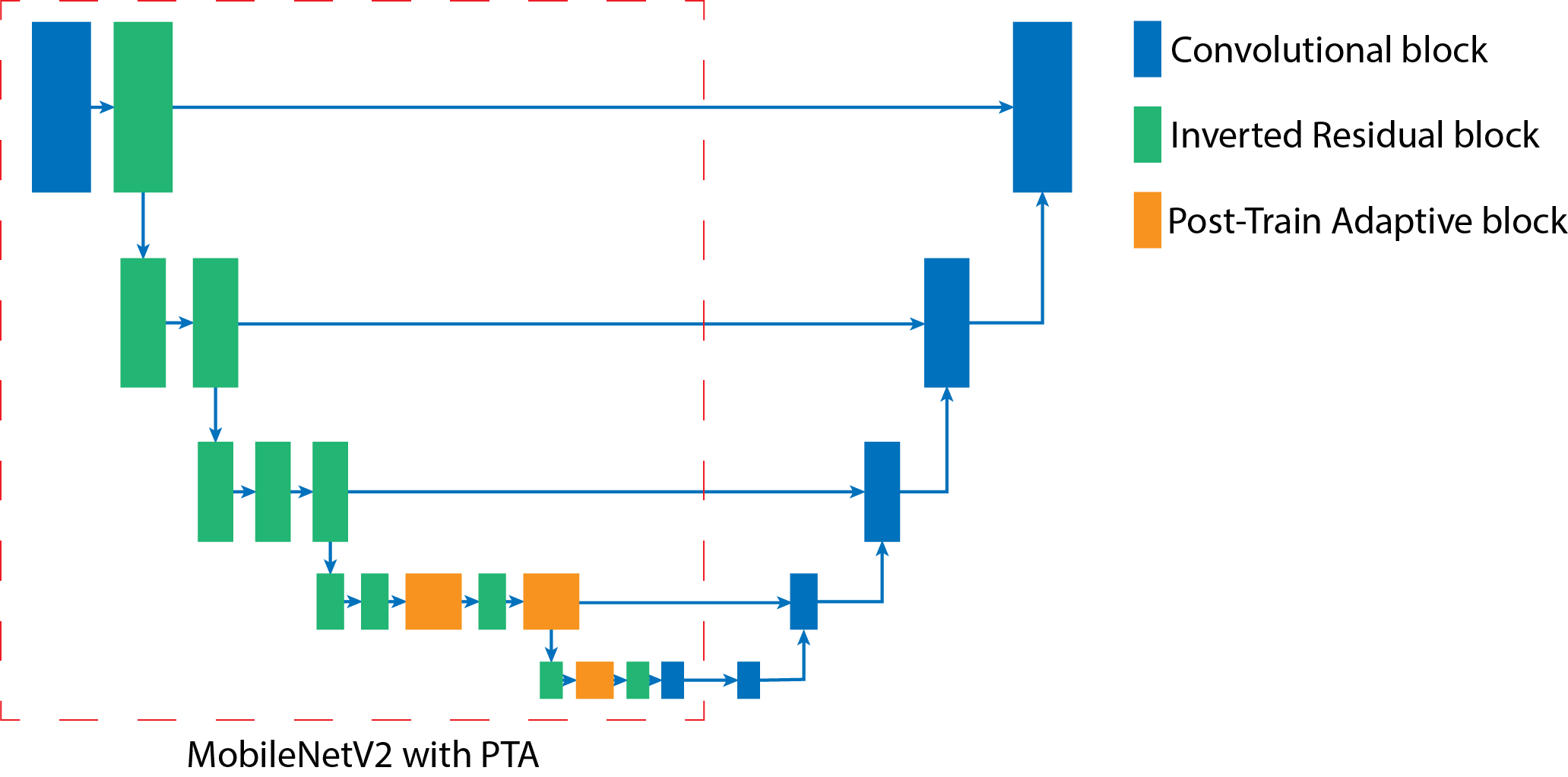}
  \caption{The proposed U-Net+PTA architecture for the image segmentation task. MobileNetV2 architecture with added PTA blocks is used as an encoder. Convolutional blocks are shown in blue, Inverted Residual in green, Post-Train Adaptive (PTA) in orange.}\label{fig:unet-mobilenetv2-pta}
\end{figure*}

We base on the U-Net architecture with MobileNetV2 backbone. To make the constructed neural network dynamic, we use the approach proposed in~\cite{PTA}, and add 3 Post-Train Adaptive (PTA) blocks to the network (as is shown in \cref{fig:unet-mobilenetv2-pta}). In this work we include PTA blocks only in the U-Net encoder (backbone), leaving the decoder part intact.

A single PTA block has 2 branches:
\begin{itemize}
  \item Light branch. Contains a single Inverted Residual block;
  \item Heavy branch. Contains two Inverted Residual blocks connected sequentially.
\end{itemize}
Each block can be dynamically configured to infer either branch exclusively, or both branches at the same time averaging the resulting feature maps.

To enable dynamic branch selection in the PTA block without retraining, a special PTA-sampling strategy is applied at training time. Specifically, several possible block configurations are selected randomly during the training procedure following the distribution shown in \cref{tab:pta-sampling}. All PTA block configurations that are possible, but not present in the table are expected to be never sampled. Note, that configuration where both blocks are enabled at the same time is also never sampled.

\begin{table}
  \caption{U-Net+PTA train-time configuration sampling. Sampling strategy follows~\cite{PTA}.}\label{tab:pta-sampling}
  \centering
  \begin{tabular}{lc}
    \toprule
    PTA Configuration     & Sampling Probability \\
    \midrule \relax
    [Heavy, Heavy, Heavy] & 0.45                 \\ \relax
    [Light, Heavy, Heavy] & 0.15                 \\ \relax
    [Heavy, Light, Heavy] & 0.15                 \\ \relax
    [Heavy, Heavy, Light] & 0.15                 \\ \relax
    [Light, Light, Light] & 0.10                 \\
    \bottomrule
  \end{tabular}
\end{table}

To train the neural network, we use the Dice\textsubscript{loss}~that has shown good segmentation training results, and to measure the resulting model quality, we use Dice\textsubscript{score}~\cite{DiceLoss}:

\begin{equation}
  \label{eq:dice-loss}
  Dice_{loss} = 1 - \frac{2\sum_i^N{p_i g_i} + \epsilon}{\sum_i^N{p_i^2} + \sum_i^N{g_i^2} + \epsilon};
\end{equation}

\begin{equation}
  \label{eq:dice-score}
  Dice_{score} = \frac{2\sum_i^N{p_i g_i} + \epsilon}{\sum_i^N{p_i^2} + \sum_i^N{g_i^2} + \epsilon};
\end{equation}
where \(p_i\) is the predicted probability distribution, \(g_i\) is the ground true one-hot vector, \(N\) is the number of classes to distinguish between, \(\epsilon\) is a small constant.

\section{Experiments}

To train and evaluate the model we use the widely known CamVid~\cite{CamVid} dataset. It contains images of size \(480 \times 360\) pixels. The dataset is split into train (367 images), validation (101 images) and test (233 images) subsets. All of the subsets have segmentation masks available of the same \(480 \times 360\) size. The task is to learn the network to segment the images into one of the following classes: sky, building, pole, road, pavement, tree, sign symbol, fence, car, pedestrian, bicyclist, unlabeled.

For training and testing we resize the images into \(256 \times 256\) size. To retain the original width to height ratio, the images are letterboxed. During training random crop and color jitter augmentations are used. Both U-Net and U-Net+PTA networks are trained for 600 epochs from scratch. No neural network pre-training is performed. Batch size is set to 8. Adam~\cite{Adam} with the learning rate of \(\alpha = 10^{- 3}\) is used as an optimizer. The results are reported on the test set. NVIDIA GTX 1050Ti is used to train and test the model. In addition, we report inference time for a batch of 8 items. To ensure accurate time measurements, timings are averaged over 1000 batches. 95\% confidence interval is given for each measurement.

\section{Results}

In \cref{tab:pta-quality} we show model performance on the test set for the original U-Net model (denoted as No PTA) and the new U-Net+PTA model (denoted as PTA-*), where * is the PTA block configuration for inference. The best result is shown in red; the second best is in blue. As is clearly seen, all PTA-based configurations show better performance that the original U-Net. The best results are obtained by PTA-HLH, followed by PTA-BBB. Note, all PTA configurations have been obtained from a single model, trained only once. Thanks to the adaptive architecture, the exact configuration can be selected after the training is complete. Interestingly, PTA-HHH, which is equivalent in architecture to the No PTA model, but has been trained with PTA-sampling strategy, is also better than No PTA configuration.

\begin{table}
  \caption{Dice\textsubscript{score} comparison for the U-Net+PTA model in the segmentation task on the CamVid dataset. Higher Dice\textsubscript{score} is better. The best configuration is highlighted in red; the second best is in blue. All Post-Train Adaptive (PTA) configurations show better quality than the original U-Net with MobileNetV2 backbone.}\label{tab:pta-quality}
  \centering
  \begin{tabular}{l@{\hspace{4em}}c}
    \toprule
    Config        & Dice\textsubscript{score} (\(\uparrow\)) \\
    \midrule
    No PTA        & 0.8583                                   \\
    PTA-HHH       & 0.8666                                   \\
    PTA-LHH       & 0.8659                                   \\
    PTA-HLH       & \best{0.8670}                            \\
    PTA-HHL       & 0.8660                                   \\
    PTA-LLL       & 0.8647                                   \\
    PTA-BBB       & \sbest{0.8667}                           \\
    \bottomrule
  \end{tabular}
\end{table}

\cref{tab:pta-inference-time} shows model complexity and inference time comparison of the U-Net and the newly proposed U-Net+PTA models. The best result is shown in red; the second best is in blue. We show post-train model configuration, the number of model parameters, the number of multiplication and addition operations for inference, absolute and relative inference time. Relative time is computed with respect to the No PTA baseline. Inference time on actual device has some fluctuation due to GPU frequency change or sporadic system activity. To ensure accurate and consistent measurements, the inference time results are averaged across 1,000 measurements. Additionally, 95\% confidence interval is given. As can be seen, PTA-based models that have one or more Light branches enabled have faster than No PTA baseline inference.

\begin{table*}
  \caption{Model complexity and inference time comparison of U-Net vs U-Net+PTA models. The following information is shown: post-train model configuration, the number of model parameters, the number of multiplication and addition operations for the inference, absolute and relative inference. Relative time is computed with respect to the No PTA model. The best result is shown in red; the second best is in blue. PTA-based models with Light branch show faster inference time.}\label{tab:pta-inference-time}
  \centering
  \begin{tabular}{lcccr}
    \toprule
    Config        & \# Params           & Multiply-Adds           & Inference Time             & Relative Impr.       \\
                  & (\(\downarrow\), M) & (\(\downarrow\), Mops.) & (\(\downarrow\), ms)       & (\%, \(\downarrow\)) \\
    \midrule
    No PTA        & 6.63                & 871.80                  & 81.37\textpm0.14           & 100.00               \\
    PTA-HHH       & 6.63                & 871.80                  & 81.16\textpm0.13           &  99.75               \\
    PTA-LHH       & 6.58                & 868.58                  & 80.21\textpm0.08           &  98.58               \\
    PTA-HLH       & 6.51                & \sbest{864.16}          & \sbest{79.78\textpm0.08}   &  \sbest{98.05}       \\
    PTA-HHL       & \sbest{6.31}        & 866.65                  & 79.89\textpm0.08           &  98.19               \\
    PTA-LLL       & \best{6.14}         & \best{855.49}           & \best{78.82\textpm0.09}    & \best{96.86}         \\
    PTA-BBB       & 7.12                & 888.12                  & 83.79\textpm0.09           & 102.98               \\
    \bottomrule
  \end{tabular}
\end{table*}

In \cref{tab:pta-training-time} we show total training time and the best Dice\textsubscript{score} for the No PTA and U-Net+PTA models. The best Dice\textsubscript{score} is selected from all possible PTA configurations from a single training pass. Both models were trained 600 epochs. As can be seen, higher Dice\textsubscript{score} for the PTA-based model is achieved with slightly faster model training.

\begin{table*}
  \caption{Training time and the best final score comparison for the U-Net and U-Net+PTA models. Higher Dice\textsubscript{score} for the PTA-based model is achieved with slightly faster model training.}\label{tab:pta-training-time}
  \centering
  \begin{tabular}{lcc}
    \toprule
    Config          & Total Training Time (\(\downarrow\), Min)     & Best Dice\textsubscript{score} (\(\uparrow\)) \\
    \midrule
    U-Net            & 161.6                                         & 0.8583                                         \\
    U-Net+PTA        & \best{158.6}                                  & \best{0.8670}                                  \\
    \bottomrule
  \end{tabular}
\end{table*}

\section{Discussion}

The Post-Train Adaptive method has been originally introduced for the task of face anti-spoofing in~\cite{PTA}. In this work we have applied it to a different computer vision task, namely image segmentation. We based our approach on the U-Net neural network with the MobileNetV2 backbone. By adding PTA blocks to the U-Net architecture and following the PTA sampling training strategy, we have been able to successfully train the neural network. The resulting network can be trained once and reconfigured later. As can be seen from the \cref{tab:pta-quality}, all of the PTA configurations show superior quality when compared to the U-Net with MobileNetV2 (No PTA). The best improvement is achieved by PTA-HLH (Dice\textsubscript{score} improvement of~0.0087), followed by PTA-BBB (+~0.0084). The PTA-HHH configuration that is equivalent in architecture to the original No PTA model is also better than the No PTA configuration, which shows the benefit of the PTA-sampling training strategy. We also note that even the lightest PTA-LLL configuration is better than No PTA baseline (+~0.0064).

The PTA-LLL configuration is the fastest configuration as is shown in \cref{tab:pta-inference-time}. PTA-LLL model shows better Dice\textsubscript{score} and is 3.14\% faster. Also, the heaviest PTA-BBB model is only 2.98\% slower, while offering 0.0084 higher Dice\textsubscript{score}. PTA-HLH has good speed and the best quality making it the best configuration in terms of speed to quality ratio.

We also note that the benefit of using 3 PTA blocks in the U-Net with MobileNetV2 backbone is smaller, than it was in the original PTA work. This can be explained by the fact that overall U-Net with MobileNetV2 backbone is a much larger model than the plain MobileNetV2 for classification as can be seen from \cref{tab:classification-vs-segmentation-pta-complexity}.

\begin{table*}
  \caption{Comparison of the number of model parameters and multiply-additions to perform classification (Class.) and segmentation (Segm.) tasks. Note, that for segmentation the baseline No PTA model is significantly larger.}\label{tab:classification-vs-segmentation-pta-complexity}
  \centering
  \begin{tabular}{lcrr}
    \toprule
    Model               & Task   & \# Params (\(\downarrow\), M) & Multiply-Adds (\(\downarrow\), Mops.) \\
    \midrule
    MobileNetV2 No PTA  & Class. & 2.23                          & 104.15                                \\
    MobileNetV2 PTA-LLL & Class. & 1.73                          &  87.84                                \\
    U-Net No PTA         & Segm.  & 6.63                          & 871.80                                \\
    U-Net PTA-LLL        & Segm.  & 6.14                          & 855.49                                \\
    \bottomrule
  \end{tabular}
\end{table*}

The PTA training procedure is easy to integrate into existing pipelines. It offers the benefits of extra model configuration after the training is complete, higher model quality, and lower inference time. In addition to that, overall U-Net+PTA training time is no larger than that of a simple U-Net model as can be seen from \cref{tab:pta-training-time}.

\section{Conclusions}

In this work Post-Train Adaptive approach has been first applied to the task of image segmentation. The PTA approach has made it possible to reconfigure the architecture of the designed neural network after the training process has been complete. The two key components of the approach are PTA blocks and PTA-sampling training strategy. The PTA blocks were added into the U-Net neural network with MobileNetV2 backbone. The post-train configuration can be done at runtime on any inference device including, but not limited to mobile devices.

In addition to post-train neural network configuration, the PTA approach has allowed to improve image segmentation quality (Dice\textsubscript{score}) on the CamVid dataset.

The final trained model can be switched at runtime between 6 PTA configurations. These configurations differ by inference time and quality. The best speed is offered by PTA-LLL configuration, that is faster and has higher quality than No PTA baseline. The best quality is achieved by PTA-HLH configuration with better than No PTA inference speed making the best configuration in term of speed to quality ratio. Importantly, all of the configurations have better quality than the original U-Net (No PTA) model.

The possible future research direction is to expand the inference time difference between heavy and light configurations to allow a single trained PTA-based network to target even more device performance categories.

\begin{small}
  \subsection*{Funding}

  The work is supported by the state budget scientific research project of Dnipro University of Technology ``Development of New Mobile Information Technologies for Person Identification and Object Classification in the Surrounding Environment'' (state registration number 0121U109787).

  \printbibliography

\end{small}

\end{document}